\documentclass{svproc}
\usepackage{graphicx}
\usepackage{amsfonts}
\usepackage{hyperref}
\usepackage{url}

\begin{document}
%
%
%
\mainmatter              
%
\title{A generic ontology and recovery protocols for Human-Robot Collaboration (HRC) systems }
\titlerunning{Human-Robot Collaboration (HRC) systems}  
%
\author{ Kamil Skarzynski\inst{2} \and   Marcin Stepniak\inst{2}  \and Waldemar Bartyna\inst{2} \and Stanislaw Ambroszkiewicz\inst{1}\inst{2}}
\authorrunning{K. Skarzynski et al.} 
%
\tocauthor{Kamil Skarzynski, Marcin Stepniak, Waldemar Bartyna, Stanis\l aw Ambroszkiewicz}
\institute{Institute of Computer Science, Polish Academy of Sciences, Jana Kazimierza 5, PL-01-248 Warsaw, Poland, 
\email{sambrosz@gmail.com},\\ 
\and
Siedlce University of Natural Sciences and Humanities, 3 Maja 54, 08-110 Siedlce, Poland, 
	\email{kamil.skar@gmail.com, martinus.st@gmail.com, wbartyna@gmail.com}
}

\maketitle  

\begin{abstract}
Recently, human–robot interactions and collaborations have become an important research topics in Robotics. Humans are considered as integral components of Human-Robot Collaboration (HRC) systems, not only as object (e.g. in health care), but also as operators and service providers in manufacturing. Sophisticated and complex tasks are to be collaboratively executed by devices (robots) and humans. 
We introduce a generic ontology for HRC systems. Description of humans is a part of the ontology. Critical and hazardous (for humans) situations, as well as corresponding safeguards are defined on the basis of the ontology. The ontology is an extension of the ontology introduced in Skarzynski et al. (2018) \cite{TAROS2018}. 
The architecture of SO-MRS \cite{TAROS2018}, a software platform for automatic task accomplishment, is extended to HRC systems.   
Ongoing experiments, carried out in a simulated HRC system,  are to verify the ontology and the architecture. 
\end{abstract}

\section{Introduction}

Physical human-robot interactions are becoming more and more important in many applications, from health care and service robotics to manufacturing. During the interactions, robots must not cause any harm to humans. 

Standards ISO 10218 and ISO/TS 15066 
 \url{https://www.iso.org/obp/ui/fr/#iso:std:iso:ts:15066:ed-1:v1:en}
provide some guidelines and requirements for designing, planing and executing human-robot collaborative tasks in the form of safeguarding for protecting the humans. 

The ISO standards are not formal and not sufficient to define generic safeguards.
No generally accepted solution have been proposed so far to implement these requirements in a universal way. Existing solutions are dedicated to specific scenarios, and can not be generalized to an ontology and protocols. 
Hence, ISO/TS 15066 may be seen only as a good and approved guidance to create a formal ontology for defining safeguards. 

IEEE Standard Ontologies for Robotics and Automation (Std 1872-2015) 
\url{https://ieeexplore.ieee.org/document/7084073} 
defines a core ontology that allows for the representation of, reasoning about, and communication of knowledge in the robotics and automation (R\&A) domain. This ontology includes generic concepts as well as their definitions, attributes, constraints, and relationships.

From Introduction of IEEE Std 1872-2015 document: 
\\ 
{\em "The growing complexity of behaviors that robots are expected to present naturally entails the use of increasingly complex knowledge as well as the need for multi-robot and human-robot collaboration. 
... Ontology plays a fundamental role in this context.''}

In IEEE Std 1872-2015 ontology, humans are needed only for  
a semi-autonomous robot, i.e.  a robot accomplishing a task in which the robot and a human operator plan and conduct the task. It requires various levels of human interaction via {\em HumanRoboCommunication}, i.e. a transfer of information between humans and robots. Hence, here the human-robot interactions are extremely limited and are not physical. 

Humans can perform operations (provide services) that are parts of a complex tasks to be accomplished in HRC systems.  Humans may be also ``objects'' of such operations in health care and rescue actions in accidents with human victims. 

We are going to define a simple and generic ontology where full human-robot interactions can be described, that is, where humans can be service providers, as well as ``objects''.  

On the basis of this ontology, we explore the possibility of the human-robot collaboration in unstructured environments challenged by frequent failures of robots and humans. 
Generic protocols for recovery from failures are a solution to this problem. If one robot (or a human) fails, another robot or a human may continue successfully the task accomplishing. 

Human presence and activity in HRC systems may cause critical and hazardous situations for humans. 
It is clear that constrains on human-robot interactions are needed in the spirit of the famous Three Laws of Robotics by Isaac Asimov. 

The constrains, corresponding safeguards, and the recovery protocols can be defined on the basis of the proposed ontology. 
They naturally extend the SO-MRS architecture (Skarzynski et al. (2018) \cite{TAROS2018}) where no humans were involved in task  accomplishing.

\section{The idea and related work}

Heterogeneous open distributed system (HODS for short) consists of  environment, and of devices (machines and robots) as well as humans that operate in the environment and may change local states of the environment. For sophisticated  tasks, a collaboration of devices and people may be necessary.  A device and a human may be considered as objects of the environment, and their state may be subject of change, e.g. their positions.   

It is supposed that the devices may be heterogeneous, and can be added to the system as well as to be removed without affecting its basic functionality, i.e. the ability for task accomplishing. There are also humans that can perform some elementary tasks. Hence, the class of the tasks is not fixed and depends on the joint capabilities of the devices and people that are currently in the system. Since such tasks can not be hard-coded in the system, there must be a language for the task specification. Intuitively, a task is an intention to change local state of the environment. That is, task consists of precondition and effect. Sometimes, the precondition is not necessary.  
Precondition specifies initial local state of the environment, whereas the effect specifies the desired local state of the  environment after the task accomplishing. Preconditions and effects are formulas in a formal language. 
So that, a formal representation of the environment (ontology) is needed. 
Task is represented by $(\phi \to \psi)$ where $\phi$ is precondition formula, and $\psi$ is effect formula. Note that here $\to$ is not the logical implication. Its meaning is to change a local state of environment. 

The principal goal of a HODS is to plan and to accomplish complex sophisticated tasks. It is clear that a generic infrastructure (in the form of software platform based on ontology and protocols) is needed. 
It is supposed (in our paper) that any HRC system is a HODS. 

IEEE Standard (Std 1872-2015) 
is quite complex. In particular, process is defined as a primitive notion {\em that happens in time and have temporal parts or stages}. 
A simple static ontology without time and processes is sufficient. The simplicity should be viewed here as an advantage. Such ontology was proposed in Skarzynski et al. (2018)  \cite{TAROS2018} for adaptive multi-robot systems without humans. 

In our approach, processes can be defined using the Service Oriented Architecture (SOA) paradigm from Information Technology, where devices and humans are represented by the services they can provide. 

\subsection{Related work}

Most of the related work has been presented in  Skarzynski et al. (2018)  \cite{TAROS2018} 
also available at arXiv \url{https://arxiv.org/abs/1709.03300}.

Let us only cite the view on the research on Multi-Robot Systems  by  Chitic, Ponage and Simonin (2014) \cite{chitic2014middlewares}: 
{\em 
``Despite many years of work in robotics, there is still a lack of established software architecture and middleware, in particular for large scale multi-robots systems. Many research teams are still writing specific hardware orientated software that is very tied to a robot. This vision makes sharing modules or extending existing code difficult. A robotic middleware should be designed to abstract the low-level hardware architecture, facilitate communication and integration of new software.''} 

The literature concerning {\em middleware} for HRC systems is rather limited. 

Lasota et al. (2017 ) \cite{Lasota} surveyed and categorized prior research addressing safety during human-robot  interaction (HRI). The authors identified four main methods of providing safety: control, motion planning, prediction, and consideration of psychological factors. They concluded that {\em  ``... ensuring safe HRI remains an open problem. Novel, robust, and generalizable safety methods are required in order to enable safe incorporation of robots into homes, offices, factories, or any other setting''.} 

Magrini et al. (2020) \cite{Magrini} proposed a specialized and dedicated safety framework to ensure coexistence of a human operator in a robotic cell in which a standard industrial robot is in motion. It is based on online monitoring of relative human–robot distance using depth sensors. 

Extensive literature reviews of the aspects of safety and failures in human-robot interactions can be find in: Honig et al. (2018)  \cite{honig2018understanding},  Robla-G{\'o}mez et al. (2017) \cite{Robla} and Villani et al. (2018)  \cite{villani2018survey} both concerning industrial environments.   

As to ontology for HRC systems, besides already mentioned IEEE Std 1872-2015, 
there are publications that are, in fact, based on this standard ontology. They emphasize the importance of the ontology for Robotics without proposing essentially new (comparing to IEEE Std 1872-2015) ontology; 
e.g. 
Fiorini et al. (2017) \cite{fiorini2017suite} and Kumar et al. (2019) \cite{kumar2019ontologies} 
to mention the most important ones.

\section{Software infrastructure (middleware) for HRC systems }

   \begin{figure}[thpb]
      \centering
      \includegraphics[width=\linewidth]{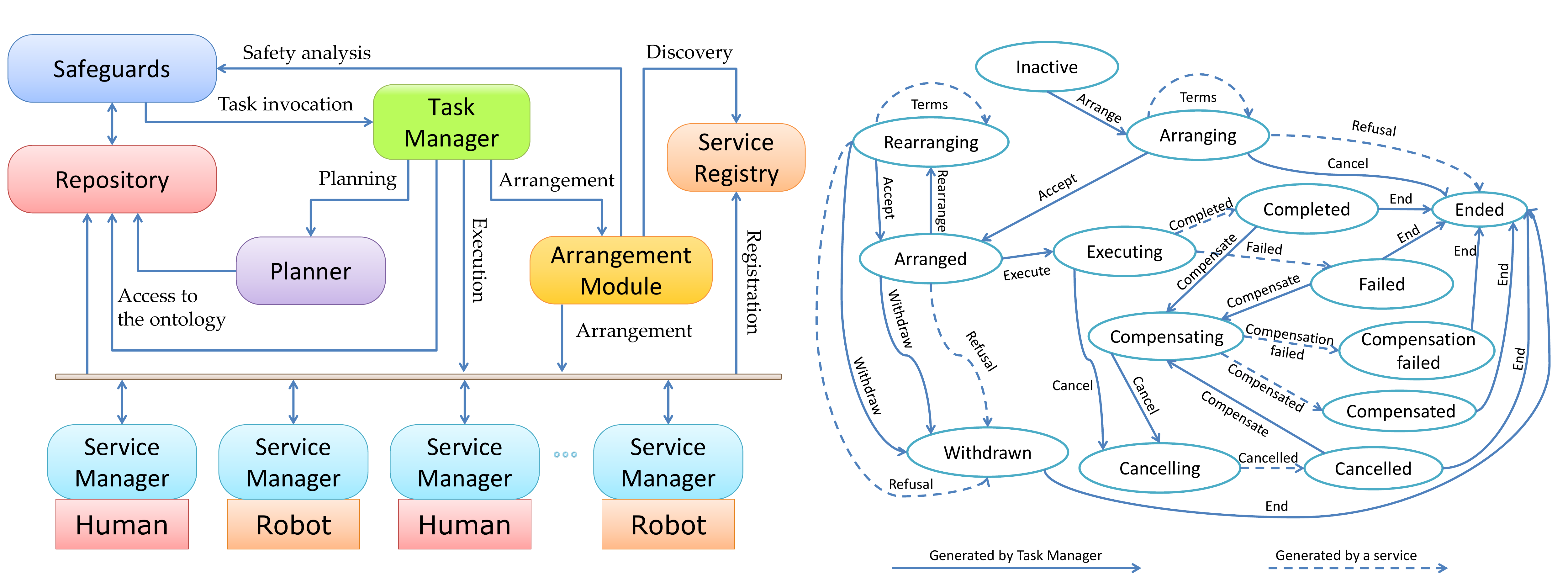}
      \caption{SO-HRCS architecture, and the state transition diagram of Failure Recovery Protocol}
      \label{protocol}\vspace{-0.3cm}
   \end{figure}

A software infrastructure is necessary to automatically accomplish complex tasks in a HRC system. An architecture of such infrastructure is presented in Fig. \ref{protocol} (left part) where Task Manager, Service Managers, Service Registry, Repository, Planer, Arrangement Module, and Safeguards are software applications that communicate using generic protocols shown in the right part of  Fig. \ref{protocol}. 

It is supposed that devices and people (providing services in a HRC system) are not isolated, i.e. there is a minimum communication in the system in the form of (wireless) network. Each device and each human (e.g. via a mobile phone) can receive and send messages. 

Devices and people are represented by their abilities (services) to change local states of the environment. 

Each device and each human is autonomous and may provide some services (via its Service Manager) to a client called Task Manager. If a Task Manager has a task to be accomplished, it sends a request (intention) to a Service Manager. Then, the corresponding service may accomplish the task, if it has enough resources and capabilities. If the service agrees, it sends (via its Service Manager) a commitment to the Task Manager. If the terms of the commitment are satisfactory for the Task Manager, it can send a request for the service invocation.  

So that, devices and humans provide some services that correspond to some types of elementary tasks they can  accomplish. The formal specification (expressed in a language of the common ontology, e.g. OWL-S \cite{OWL}) of the type of a service consists of a precondition and an effect. 

Note that service type has the same syntactical form as task. 

The type of service (provided by a device or a human) must be published (via Service Manager of a device or a human) to a Service  Registry. A Task Manager may discover the service in Service Registry, and invoke it. This constitutes the essence of the SOA paradigm applied to HRC systems. 

%
Repository (the next component of the middleware for HRC system) is a realization of the common knowledge of the environment representation (ontology, see Section \ref{ontology}), and a storage of the current maps of the environment, i.e. instances of the ontology. Since the environment may be changed by devices and people, the maps must be updated. There are   also Planer, and Arrangement Module. 

If a Task Manager wants to realize a complex task (a sequence or partial order of elementary tasks), then some services, that may jointly accomplish the complex task, should be discovered (in a Service Registry), and composed into a workflow (composite service) via the Arrangement Module. 

Task Manager is responsible for constructing an abstract plan (via the  Planner) in the form of partial order of service types. Then, appropriate concrete services should be arranged into a concrete plan (workflow) via the Arrangement Module. Finally, the workflow is executed and its performance is monitored. If a failure occurs (due to a broken communication or inability of a service to fulfill the arranged commitment), then failure recovery mechanisms must be applied. 
Simple mechanisms (in the form of protocol, see right part of Fig. \ref{protocol}) consist in re-planning, and  changing some parts of the workflow in order to continue the task execution. This constitutes the SO-MRS architecture proposed in Skarzynski et al. (2018) \cite{TAROS2018}. 

Since people are involved in  HRC systems, additional failures related to unexpected events (i.e. critical situations where human safety is at risk) must be introduced in order to protect the people. 
For this very reason an additional component of the software infrastructure is necessary. It is Safeguards where the critical situations are defined. 

Critical situations are taken into account by Planner when an abstract plan is needed for a task. Any abstract plan and concrete plan (workflow) should exclude evident critical situations.  Generally,  occurrences of critical situations,  during task accomplishing, can not be excluded at the level of planning. It is reasonable to introduce a mechanism for risk assessment in the planning to reduce the occurrences as much as possible.  

Each process of a task accomplishing (supervised by a Task Manager), and each individual service execution (supervised by an appropriate device or a human) must react on the critical situations. The reaction (called {\em safeguard}) should be of the following form. If a {\em critical situation} occurs, then an action is executed that transform the critical situation into a {\em safe situation}. 

A critical (hazardous) local situation as well as the corresponding safe situation should be defined as formulas in the language of ontology. 

Generally, situation is defined as a formula, and identified with the set of local states satisfying the formula.  

Let a formula defining a critical situation (resp. safe situation)  be called a {\em  critical formula} (resp. {\em safe formula}). 

Some critical situations (formulas) and safe situations (formulas) are universal and independent of concrete applications.  


{\bf Definition.} 
Let $\phi$ denote a critical situation, and $\psi_1$, $\psi_2$, ... $\psi_n$ be a finite sequence of safe formulas. Then, {\em safeguard} is defined as $\phi \to (\psi_1, \psi_2, ...  \psi_n)$. 
 
The meaning of a safeguard is as follows. If $\phi$ is true in the current local state of the environment, then change the state to a safe state where  at least one of the formulas $\psi_1$, $\psi_2$, ... $\psi_n$ is true in that state. The change must be possible, so that one of the tasks 
$(\phi \to \psi_1)$, $(\phi \to \psi_2)$, ... $(\phi \to \psi_n)$ can be accomplished successfully, either by a single service or by a composite service invoked by the appropriate Task Manager.  
Actually, it is a strong condition. 

A task is a safeguard, if the task precondition is a critical formula, and the task effect is a safe formula. 

Safeguards are stored in the Safeguards module. 

Introduction of safeguards requires an extension of: the ontology of SO-MRS, functionalities of Task Manager and Service Manager, and corresponding protocols. 
Task Manager must take into account the critical situation in the planning. 

A service provider (a device or a human) must monitor the local environment during its service execution. If a critical situation occurs, then the service provider is obliged to report this to the Task Manager, and take appropriate means (if possible) to change the critical situation to a safe one. 

If the service provider can not manage, it means {\em a critical failure} that must be reported to the Task Manager that is responsible for a recovery from the failure, i.e. for a transformation to a safe local state, if possible. A safeguard stored in Safeguards module, is used by the Task Manager as a task to be accomplished in order to change the critical situation into a safe one. 

To summarize, the proposed software infrastructure of HRC system (for automatic complex task accomplishing) consists of services (represented by Service Managers) provided by devices and humans, Service Registry, Task Manager (with Planner and Arrangement Module), Safeguards, and Repository of the current maps (instances of ontology) of the environment. The interactions between them are based on generic protocols for publishing, discovering, composing elementary services,  arranging, execution, monitoring and recovery from failures. 

Note that ontology is the basis for the protocols. It allows to specify local states of the environment, tasks, service types,  intentions, commitments, critical situations for humans, and situations resulting from failures. The protocols and the ontology determine the new architecture called Service Oriented Human-Robot Collaboration System (SO-HRCS for short).  SO-HRCS is a substantial extension of SO-MRS (Service Oriented Multi-Robot System) architecture introduced in Skarzynski et al. (2018) \cite{TAROS2018}. 

SO-HRCS architecture allows several independent Task Managers, Service Registries, and Repositories. Note that the presented approach is at higher level of abstraction than Robot Operating System (ROS) that is usually used to implement services on the devices. 

The main contribution of the paper consists of a simple universal upper ontology describing HRC systems, SO-HRCS architecture, and some new  protocols related to the new module, i.e. Safeguards. Since an occurrence of a critical situation is considered as a failure, the protocol for failure handling and recovery from SO-MRS is adopted for SO-HRCS. The protocol is briefly presented in the next subsection.

\subsection{Protocol for failure handling and recovery }
 
Since some ideas and methods are adopted from electronic business transactions, realization of a task is called {\em a transaction}. Participants of a transaction are the services involved in the corresponding task accomplishing.  

A transaction is successfully completed, if its task is accomplished.
The transaction mechanism designed for handling failures has the following properties. 
\begin{enumerate}
  \item Failed services may be replaced by other services during task accomplishing. 
  \item General plan may be changed.
  \item Transaction ends either after successful completion of the task, or inability to complete the task, or cancellation of the task.
\end{enumerate}

In distributed systems, a communication protocol specifies  format of messages exchanged between two or more communicating parties, message order, and actions taken when a message is sent or received. Based on the OASIS Web Services Transaction (WS-TX 1.2) standards (2009)  \url{https://www.oasis-open.org/committees/ws-tx/}, 
a transaction protocol, called  Failure Recovery Protocol (FRP, for short) proposed in Skarzynski et al. (2018) \cite{TAROS2018}, is adopted for HRC systems. FRP defines states of services, and types of messages exchanged between Task Manager and services, see Fig.~\ref{protocol}.  

 Task Manager uses FRP to initialize particular phases of service invocation, monitoring their progress, and take  additional actions, e.g.  compensations. 
Task Manager invokes a service by sending the input data (specified in the commitment) to the service via its Service Manager. The service sends messages (via its Service Manager and according to the protocol) to notify Task Manager about the performance of the delegated sub-task specified in the commitment.

 After successful execution, the service (via its Service Manager) sends to the Task Manager the confirmation of sub-task completion, e.g. changing local situation in the environment to the one specified in the commitment. Task Manager can also stop the service execution before its completion. This may be caused by the task cancellation by the client, a failure during execution of other services in the plan (that cannot be replaced), or by changes in the environment making the current plan infeasible. 

A robot (or a human as a service provider) may not be able to successfully complete a sub-task. In this case, its Service Manager notifies the Task Manager by sending a detailed description of the problem. On this basis, the Task Manager can take appropriate actions. If a Service Manager is not able to send such information, the Task Manager must invoke appropriate cognitive service (a patrolling robot or a human, if available) to recognize the situation resulting from the failure.  

Compensation is performed either after a cancellation of a sub-task execution by a service, or after the occurrence of a failure that interrupts the execution. It is designed to restore the original state of the environment before the execution. Since restoring that situation is sometimes impossible, the compensation may change the situation,  resulting from the failure, to a situation from which the task realization can be continued. Note that even for simple transportation tasks (that seem to be simple) a universal failure recovery mechanism and corresponding compensations are not easy to design and implement. 
A concrete plan should contain  predefined procedures for failure handling and compensations. 

Note that in the protocol, humans and devices are viewed only as service providers.  
Critical situations, that occur during sub-task execution by a service, and cannot be managed by the service, are considered as failures in the protocol.

\subsection{Services}


There are the following three kinds of services in HRC systems:   
\begin{enumerate}
\item 
Physical services that may change local situations in the physical environment.  
\item 
Cognitive services that can recognize situations described by formulas of the language of the ontology.  
\item 
Software services that process data.
\end{enumerate}
Physical services and cognitive services can be performed by humans. 

Complete service description consists of the following elements:
\begin{itemize}
\item Name of the type of service, i.e. name of an action that the service performs. 
\item Specification of the inputs and outputs of the service. 
\item The condition required for service invocation (precondition), and the effect of service invocation. 
\item Service attributes as information about the static features of a service, e.g. operation range, cost, and average realization time.
\end{itemize}

Precondition and effect are defined as formulas of a formal language (e.g. OWL,
\cite{OWL}
  or Entish~\cite{ambroszkiewicz2004entish}) describing local situations in the environment.  
Entish is a simplified version (without quantifiers) of the first order logic.  It has logical operators (\emph{and}, \emph{or}), names of relations (e.g., \emph{isIn}, \emph{isAdjacentTo}), names of functions (e.g., \emph{action}, \emph{range}), and variables.  A precondition formula is a description of the initial situation, and the effect formula is a description of the desired final situation.

Concrete service providers are devices, sensors, humans, and computers (servers). 
Each of them may provide several different services.

\section{Ontology for HRC systems}
\label{ontology}

Upper ontology is  {\em a general structure of the representation of environment of a HRC system}.  It is a formal and abstract description of concepts (objects) and relations between them.  
A concrete model of the environment of a HRC system (called a map) is an instance of the upper ontology, where the objects and the relations are specified. 

The upper ontology is based on the following general concepts:
\begin{itemize}
\item Attributes of objects,  e.g.: color, weight, volume, position, rotation, shape, texture, etc. They are recognizable and measurable by devices and/or humans in HRC system. 
\item Predefined relations on attributes. 
\item Relations between objects are defined on the basis of the predefined relations. 
\item Type of object is uniquely determined by the object construction. 
\item Physical type is defined only by some attributes and relations between them. 
\item Abstract type is defined by already defined types (physical  and/or abstract), and relations between objects of these types. For example, the abstract type \emph{Building} consists of several other abstract types like: storey, passages, rooms, stairs, lifts, etc..  Internal structure of an object of  type rooms is composed of objects of physical types such as walls, floor, ceiling, windows, and doors, as well as the relations between these objects. 
\item Object is an instance of its type with concrete attribute values, concrete sub-objects, and concrete relations hold between attributes, and between the sub-objects.
\end{itemize}
\begin{figure}[htb]
\includegraphics[width=0.99\linewidth]{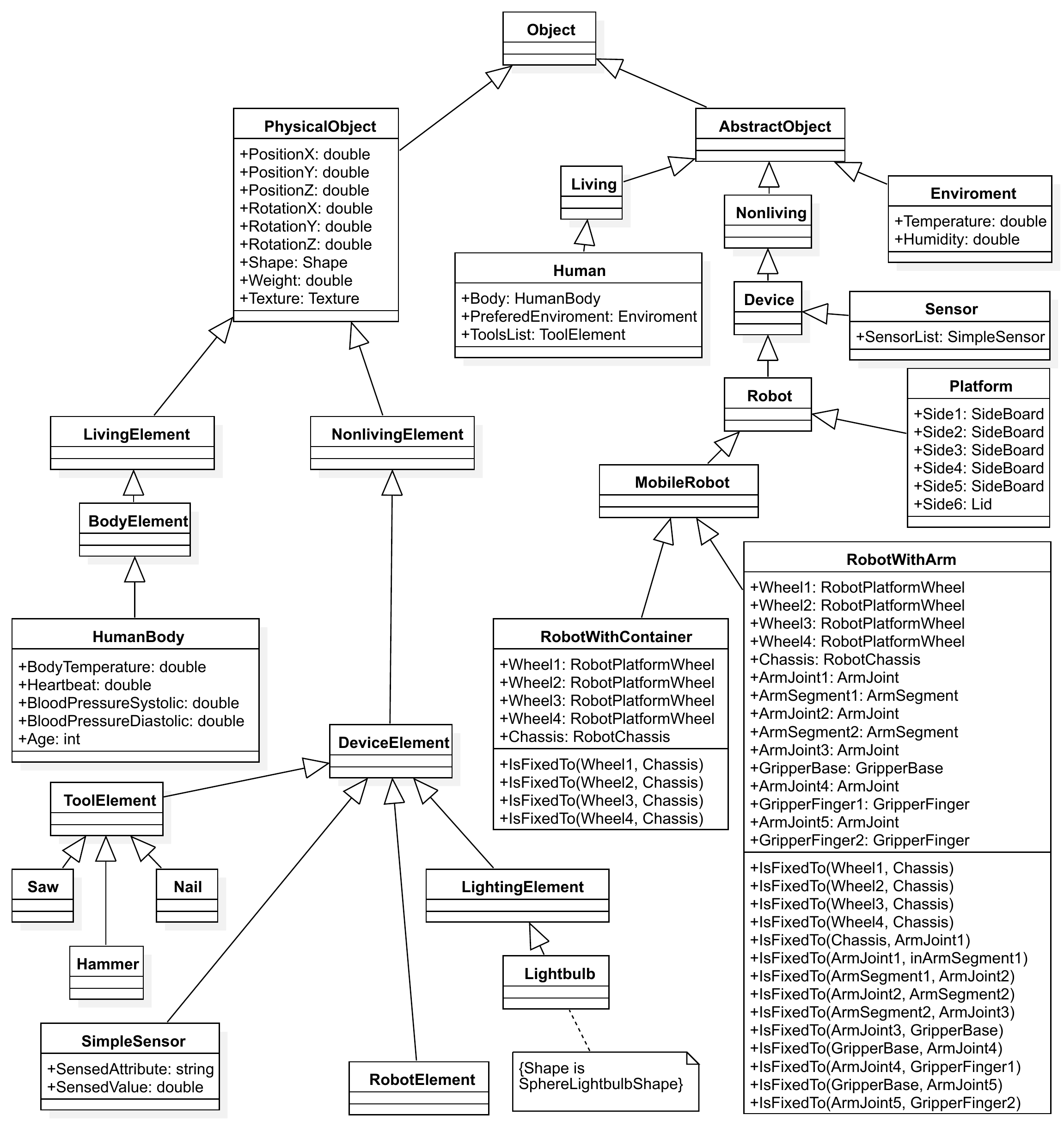}
\caption{A part of the upper ontology for HRC systems}
\label{fig:ontology}\vspace{-0.2cm}
\end{figure}

Upper ontology is defined as a hierarchy of types in the form of tree. Parent type - child type relation in the tree means inheritance, i.e. child inherits from its parent. The root of the tree is \emph{Object}. It has two main branches starting with \emph{PhysicalObject} and \emph{AbstractObject} from the root, see  Fig. \ref{fig:ontology}. These two types separate physical objects (directly recognizable by robots and humans) from abstract objects that are hierarchically composed from physical objects, relations between them, and attributes.  

The physical types are leaves of the \emph{PhysicalObject} branch. The abstract types are leaves of the \emph{AbstractObject} branch. 

Intermediate types (nodes of the tree that are not leaves in both branches) serve for inheritance and taxonomy. 

In order to add a new type to the ontology one has to specify:
\begin{itemize}
\item Parent type, i.e. the type that the new type inherits from. 
\item List of additional attributes of the new type.  
\item List of types of obligatory sub-objects, i.e. types of objects that are integral parts of the type being defined, e.g. legs in the case of the type of table. 
\item List of constraints and relations for the attributes. List  of obligatory relations between sub-objects.
\end{itemize} 

Let us again stress that any physical type is defined as a collection of attributes with restricted ranges, and relations between them. Any abstract type is defined recursively from physical types. 
Hence, the attributes and predefined relations between them are the basic elements for construction the upper ontology.  

A particular object (as an instance of its type) is defined by specifying concrete values of its attributes satisfying the relations, and specifying its sub-objects (if it is of abstract type) that satisfy the relations between them. 

The upper ontology is a hierarchy of types. An instance of the ontology is called a map of the environment. It is important that in order to support an automatic map creating  and updating (by mobile robots and sensor networks, also by humans), the attributes must be recognizable and measurable by robot and/or human sensors. 

The original upper ontology proposed in Skarzynski et al. (2018) \cite{TAROS2018} is extended here by two branches of the hierarchy tree, see Fig. \ref{fig:ontology}. The first one (called \emph{LivingElement}) is for \emph{PhysicalObject} node. The second one (called \emph{Living}) is for \emph{AbstractObject} node.

\emph{PhysicalObject} type is inherited by \emph{NonlivingElement} and \emph{LivingElement}.  

\emph{HumanBody} is an elementary type that inherits from \emph{BodyElement} type that, in turns, inherits from \emph{LivingElement} type. 

One can imagine other types that inherit from  \emph{LivingElement} type. 
\emph{BodyElement} type may be inherited by, for example,  \emph{DogBody} type and \emph{HorseBody} type yet to be defined. 

\emph{HumanBody} type is composed of such basic attributes as: body temperature, heart rate and blood pressure, etc.  that must be defined first. 

\emph{NonlivingElement} type is inherited by \emph{DeviceElement} that is, in turn, inherited by \emph{RobotElement}, \emph{ToolElement}, \emph{SimpleSensor}  and \emph{LightingElement}. 

\emph{SimpleSensor} is designed to sense and store information about a specific place in the environment. It it composed of the list of pairs: {SensedAttribute} (e.g. humidity, light intensity, temperature, CO pollution), and its value.   
It is used to define abstract \emph{Sensor} devices by aggregating individual sensors, e.g. a weather station (for measuring temperature, humidity, light intensity, etc.) to recognize the current local state of the environment.

All these physical types are components for creating abstract   object types. 

\emph{AbstractObject} type is inherited by 
two types \emph{Living} and \emph{Nonliving} that are dedicated to service providers in HRC systems. 

\emph{Human}, that inherits from \emph{Living} type, is defined by the following three attributes.  \emph{Body} value is an element of elementary physical type \emph{Human}. 
\emph{ToolList} is a list of {\em tools} that the Human can operate with.  \emph{PreferredEnvironment} is an element of abstract type \emph{Environment} and defines non critical conditions for the Human, like temperature, humility, and radiation. Perhaps some additional attributes are needed for complex scenarios to be realized in HRC systems, and for the roles of humans in that scenarios. 

\emph{Nonliving} type is inherited by \emph{Device} that is, in turn, inherited by \emph{Robot} and \emph{Sensor}. 

\emph{Robot} is inherited by \emph{MobileRobot}, \emph{RobotWithArm}, \emph{Platform}, and \emph{RobotWithContainer} that specify concrete types of robots composed of concrete physical elementary objects of type \emph{RobotElement}. 

The types shown in Fig. \ref{fig:ontology} are self explained by their names. Actually, for the sake of presentation, it is only a small part of the proposed upper ontology for HRC systems.   

The ontology is still under construction. Although the main types, physical and abstract ones, seem to to have sense, a lot of  research is needed to develop types that adequately represent real environment where robots and humans can safely collaborate in accomplishing sophisticated tasks.

\subsection{Conclusions }

The presented work is a continuation of the project RobREx RobREx - Autonomy in rescue and exploration robots  (2012-2015) PBS1/A3/8/2012  NCBiR) with new versions of the ontology and new protocols; see \url{http://www.robrex.ipipan.eu/about.php?lang=en} for the experiments. 

The proposed SO-HRCS architecture is under ongoing testing and evaluation in a universal simulation environment implemented in Unity 3D, and generated {\em automatically (!)} from the contents of the Repository. 

Actually, the research on the upper ontology and SO-HRCS architecture is at preliminary and experimental stage. Its value and importance for Robotics should be discussed and verified in real world applications. 

\bibliographystyle{splncs}
\bibliography{references}

\begin{thebibliography}{10}

\bibitem{TAROS2018}
Skarzynski, K., Stepniak, M., Bartyna, W., Ambroszkiewicz, S.:
\newblock {SO-MRS: a multi-robot system architecture based on the SOA paradigm
  and ontology}.
\newblock In: Annual Conference Towards Autonomous Robotic Systems, Springer
  (2018)  330--342 also available at arXiv
  \url{https://arxiv.org/abs/1709.03300}.

\bibitem{chitic2014middlewares}
Chitic, S.G., Ponge, J., Simonin, O.:
\newblock Are middlewares ready for multi-robots systems?
\newblock In: International Conference on Simulation, Modeling, and Programming
  for Autonomous Robots, Springer (2014)  279--290

\bibitem{Lasota}
Lasota, P.A., Fong, T., Shah, J.A.,  et~al.:
\newblock A survey of methods for safe human-robot interaction.
\newblock Foundations and Trends{\textregistered} in Robotics \textbf{5}(4)
  (2017)  261--349

\bibitem{Magrini}
Magrini, E., Ferraguti, F., Ronga, A.J., Pini, F., De~Luca, A., Leali, F.:
\newblock Human-robot coexistence and interaction in open industrial cells.
\newblock Robotics and Computer-Integrated Manufacturing \textbf{61} (2020)
  101846

\bibitem{honig2018understanding}
Honig, S., Oron-Gilad, T.:
\newblock Understanding and resolving failures in human-robot interaction:
  Literature review and model development.
\newblock Frontiers in psychology \textbf{9} (2018)  861

\bibitem{Robla}
Robla-G{\'o}mez, S., Becerra, V.M., Llata, J.R., Gonzalez-Sarabia, E.,
  Torre-Ferrero, C., Perez-Oria, J.:
\newblock Working together: A review on safe human-robot collaboration in
  industrial environments.
\newblock IEEE Access \textbf{5} (2017)  26754--26773

\bibitem{villani2018survey}
Villani, V., Pini, F., Leali, F., Secchi, C.:
\newblock Survey on human--robot collaboration in industrial settings: Safety,
  intuitive interfaces and applications.
\newblock Mechatronics \textbf{55} (2018)  248--266

\bibitem{fiorini2017suite}
Fiorini, S.R., Bermejo-Alonso, J., Gon{\c{c}}alves, P., de~Freitas, E.P.,
  Alarcos, A.O., Olszewska, J.I., Prestes, E., Schlenoff, C., Ragavan, S.V.,
  Redfield, S.,  et~al.:
\newblock A suite of ontologies for robotics and automation [industrial
  activities].
\newblock IEEE Robotics \& Automation Magazine \textbf{24}(1) (2017)  8--11

\bibitem{kumar2019ontologies}
Kumar, V.R.S., Khamis, A., Fiorini, S., Carbonera, J.L., Alarcos, A.O., Habib,
  M., Goncalves, P., Li, H., Olszewska, J.I.:
\newblock Ontologies for industry 4.0.
\newblock The Knowledge Engineering Review \textbf{34} (2019)

\bibitem{OWL}
McGuinness, D.L., van Harmelen, F.:
\newblock {OWL Web Ontology Language Overview.
  http://www.w3.org/TR/owl-features/, February 2004. World Wide Web Consortium
  (W3C) Recommendation. }.
\newblock \url{}

\bibitem{ambroszkiewicz2004entish}
Ambroszkiewicz, S.:
\newblock Entish: A language for describing data processing in open distributed
  systems.
\newblock Fundamenta Informaticae \textbf{60}(1-4) (2004)  41--66

\end{thebibliography}

\end{document}